\pdfoutput=1

\documentclass[11pt]{article}

\usepackage[]{ACL2023}

\usepackage{times}
\usepackage{latexsym}
\usepackage{graphicx}

\usepackage{booktabs}
\usepackage{amsmath}
\usepackage{amssymb}
\usepackage{enumitem} 
\usepackage[T1]{fontenc}
\usepackage{tcolorbox}

\usepackage[utf8]{inputenc}

\usepackage{microtype}

\usepackage{inconsolata}
\usepackage[linesnumbered,ruled,vlined]{algorithm2e}
\newcommand{\rparagraph}[1]{\vspace{1.2mm}\noindent\textbf{#1}}

\newcommand{\eg}{\textit{e}.\textit{g}.,\ }

\title{Segment-Level Diffusion: A Framework for Controllable Long-Form Generation with Diffusion Language Models}

\author{
Xiaochen Zhu,
Georgi Karadzhov,
Chenxi Whitehouse,
Andreas Vlachos\\
  University of Cambridge \\
  \texttt{\{xz479, gmk34, cj507, av308\}@cam.ac.uk}
}

\begin{document}
\maketitle
\begin{abstract}
Diffusion models have shown promise in text generation, but often struggle with generating long, coherent, and contextually accurate text. 
Token-level diffusion doesn't model word-order dependencies explicitly and operates on short, fixed output windows, while passage-level diffusion struggles with learning robust representations for long-form text. To address these challenges, we propose Segment-Level Diffusion (SLD), a framework that enhances diffusion-based text generation through text segmentation, robust representation training with adversarial and contrastive learning, and improved latent-space guidance. By segmenting long-form outputs into multiple latent representations and decoding them with an autoregressive decoder, SLD simplifies diffusion predictions and improves scalability. Experiments on four datasets demonstrate that, when compared to other diffusion and autoregressive baselines SLD achieves competitive or superior fluency, coherence, and contextual compatibility in automatic and human evaluations. \footnote{ Our code is available at: \url{https://github.com/SpaceHunterInf/Segment_Level_Diffusion}}
\end{abstract}

\section{Introduction}
\begin{figure}[ht]
    \centering
    \includegraphics[width=0.48 \textwidth]{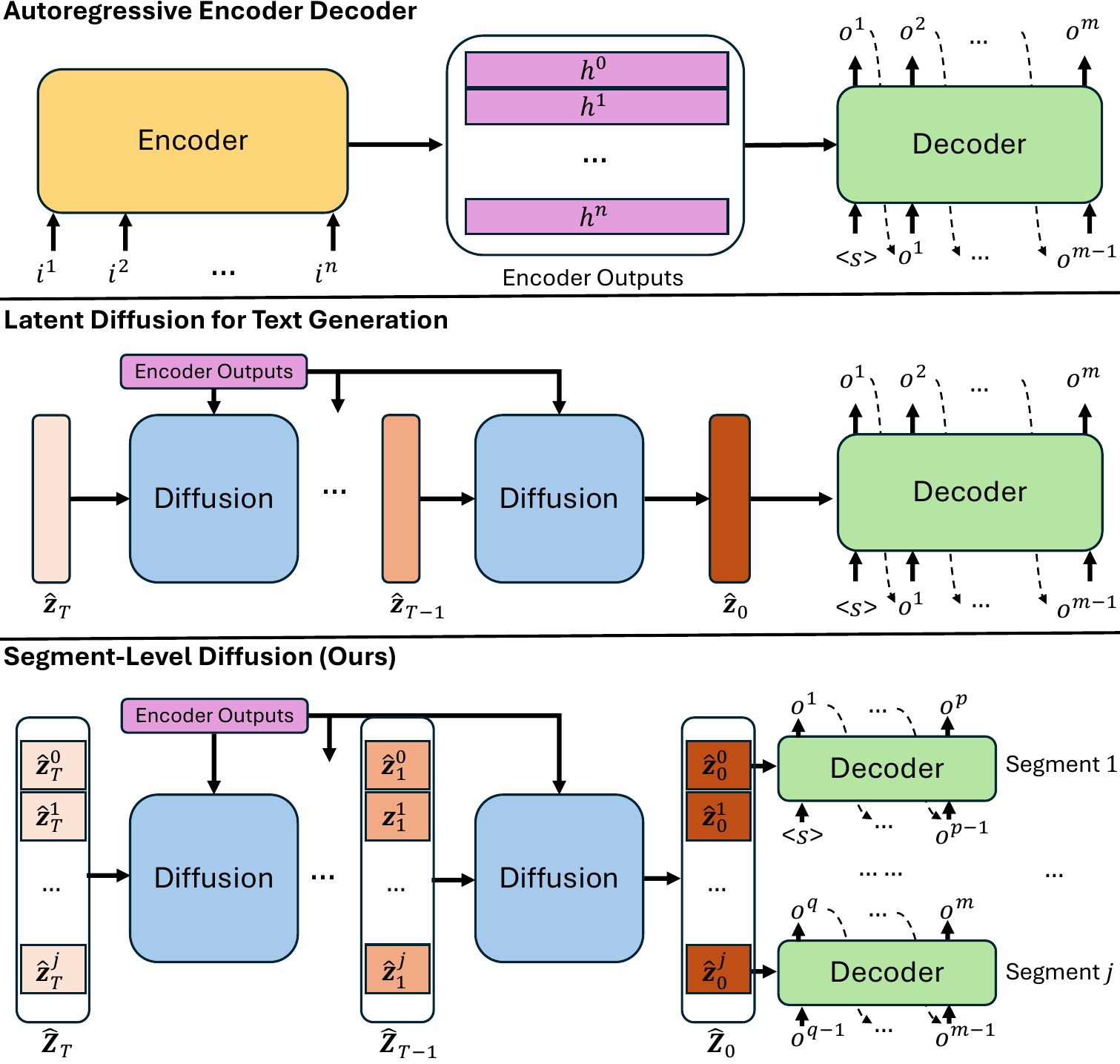}
    \caption{Comparison of AR models (top), latent diffusion (middle), and our segment-level diffusion (bottom). Unlike latent diffusion, which de-noises a single latent representation, our method splits outputs and representation into segments as the cross-attention target for conditional generation with parallel autoregressive decoding, improving text quality and controllability. }
    \label{fig:model_comparison}
    \vspace{-2mm}
\end{figure}
Transformer-based autoregressive (AR) language models have become the prevailing standard in natural language generation
\cite{vaswani2017attention, zhao2023survey}. However, the nature of next-token prediction inherently makes them prone to error propagation and incorrect handling of long-term dependencies, while also complicating controllable generation \cite{he-etal-2021-exposure, wu-etal-2018-beyond}. 

Diffusion models, which are non-autoregressive (NAR) generative models widely successful in image and video generation, have also shown promise in text generation \cite{ho2020denoising, radford2021learning, singer2022make}. \citet{li2022diffusion} pioneered the application of diffusion models to discrete text generation by predicting continuous word embeddings. Building on this work, \citet{lin2023text} introduced GENIE, a pre-trained diffusion language model that enhances semantic understanding 
through continuous paragraph-level de-noising. These approaches fall under token-level diffusion, as they directly generate word embeddings. In contrast, \citet{lovelace2024latent} proposed latent diffusion for text generation (LD4LG), encoding text into latent representations, applying diffusion to high-level semantic structures, and decoding them into text using an AR decoder. For dialogues datasets with complex intent and actions \cite{chen-etal-2021-dialogsum, hu2023multi, gliwa2019samsum}, \citet{chen2023controllable} uses diffusion models for controllable dialogue generation, operating on high-level discourse representations to enable precise control over the semantics of generated dialogues.

However, existing diffusion language models face challenges in generating longer texts \cite{yang2023diffusion, zou2023survey}. In such cases token-level diffusion becomes computationally expensive, as it either
requires pre-training with larger output windows,  or relies on iterative
diffusion sampling \cite{ tang2023can, yi2024diffusion}. The fixed output window is also wasteful when the generated sequence is shorter. Additionally, unlike AR methods, they do not model word-order dependencies explicitly, often resulting in ungrammatical or incoherent output. Generating  latent representations for passages with multiple sentences is harder, since they are highly sensitive to noise which can lead to abrupt  changes of meaning in the decoded text, and learning a smooth latent distribution is challenging \cite{vahdat2021score, zhang2024planner}.

To address these limitations, we propose a novel approach for diffusion-based text generation, Segment-Level Diffusion (SLD), illustrated in \autoref{fig:model_comparison}. 
Inspired by the concept of image patches \cite{ding2023patched},
we use a diffusion model to perform high-level semantics and structural planning, generating 
a latent representation for each 
segment (\eg sentences in paragraphs, utterances in dialogues), instead of handling long texts with a single latent representation.
Then, an AR decoder decodes predicted representations to texts. 
To improve text generation, we integrate adversarial training \citep{miyato2016adversarial} and contrastive learning \citep{gao-etal-2021-simcse} to smoothen the latent representation distribution and optimize the AR decoder with respect to the 
diffusion process. 

We compare our SLD model against three diffusion models, GENIE \cite{lin2023text}, LD4LG \cite{lovelace2024latent}, Diffuse-CG \cite{chen2023controllable}, and an autoregressive baseline, Flan-T5 \cite{chung2024scaling}. 
The evaluation includes summarization (XSum, \citealt{narayan-etal-2018-dont}), title-to-story generation (ROCStories, \citealt{mostafazadeh-etal-2016-corpus}), summary-to-dialogue generation (DialogSum, \citealt{chen-etal-2021-dialogsum}), and multiparty decision-making dialogue generation (DeliData, \citealt{karadzhov2023delidata}). 
Evaluation by both automatic and human metrics shows that SLD generates text that is more coherent and fluent, better aligned with the provided input, and matches ground-truth 
 references more closely.
\section{Related Work}

\rparagraph{Token-Level Diffusion}
\citet{li2022diffusion} adapted diffusion model for discrete text generation by operating in the continuous space of word embeddings jointly learned by the model. The architecture iteratively de-noises sampled Gaussian noise into a sequence of word vectors. A rounding method is then applied to project the  embeddings predicted into the nearest embeddings. Extending this work, \citet{gong2022diffuseq} applied token-level diffusion to sequence-to-sequence generation tasks. \citet{lin2023text} advanced this approach by incorporating pre-training, which enhanced semantic and syntactic coherence by training diffusion decoders to reconstruct clean paragraphs from corrupted embeddings.  
These models achieve sequence-to-sequence generation using encoded text as classifier-free guidance \cite{ho2022classifier}. \citet{zhou2023diffusion} unifies discrete text generation and continuous representations by using BART \cite{lewis2019bart} with self-prompting to recover masked tokens.

However, token-level diffusion has notable limitations. Unlike AR decoding methods that always condition on previously decoded tokens,  NAR generation  does not model word-order dependencies explicitly, often resulting in text that lacks grammatical correctness and fluency. Furthermore, the fixed output window restricts the length of the generated text. It is computationally expensive to re-train the entire token-level diffusion model with larger output windows, even more so for architectures without a pre-trained language model backbone \cite{gulrajani2024likelihood, loudiscrete, austin2021structured}. Even though existing literature has  accelerated diffusion sampling \cite{gong2023diffuseq, tang2023can}, token-level diffusion remains inefficient if the generated sequence is shorter than the output window, as NAR decoding always generates the full output. 

\rparagraph{Passage-Level Diffusion}
\citet{lovelace2024latent} built on the concept of latent space diffusion \cite{rombach2022high} by compressing and predicting texts using high-level semantic representations, rather than directly predicting fine-grained token representations. Such compression is beneficial for both performance and efficiency, as it provides a length-independent representation and removes information not needed for diffusion prediction, in contrast to representations from traditional language encoders. A separate AR decoder is employed to ensure the fluency of the generated text. 

However, this approach primarily focuses on short text generation, as learning robust latent representations for long passages remains challenging, and it is crucial to ensure the smoothness of the learned distribution for high-quality generation \cite{vahdat2021score}. Without proper regularization, the learned distribution may be susceptible to abrupt semantic changes due to small perturbations, increasing the difficulty of the task for the diffusion model. Although \citet{zhang2024planner} proposed techniques to improve the distributional smoothness of latent representations, the correspondence between latent representations and specific components of the generated text remains unclear. This ambiguity complicates fine-grained guidance and limits control over the generation process.

These limitations result in existing token- and passage-level diffusion models struggling to generate long and coherent text. Despite their ability to generate outputs up to 64 tokens in length, they
were primarily evaluated on tasks involving short text generation (\eg QQP paraphrasing, XSum summarization) with outputs typically around 30 tokens or less \cite{gong2022diffuseq, sharma2019natural, yi2024diffusion, li2023diffusion}.

\section{Segment-Level Diffusion}

\begin{figure*}[!htp]
    \centering
    \includegraphics[width=0.96 \textwidth]{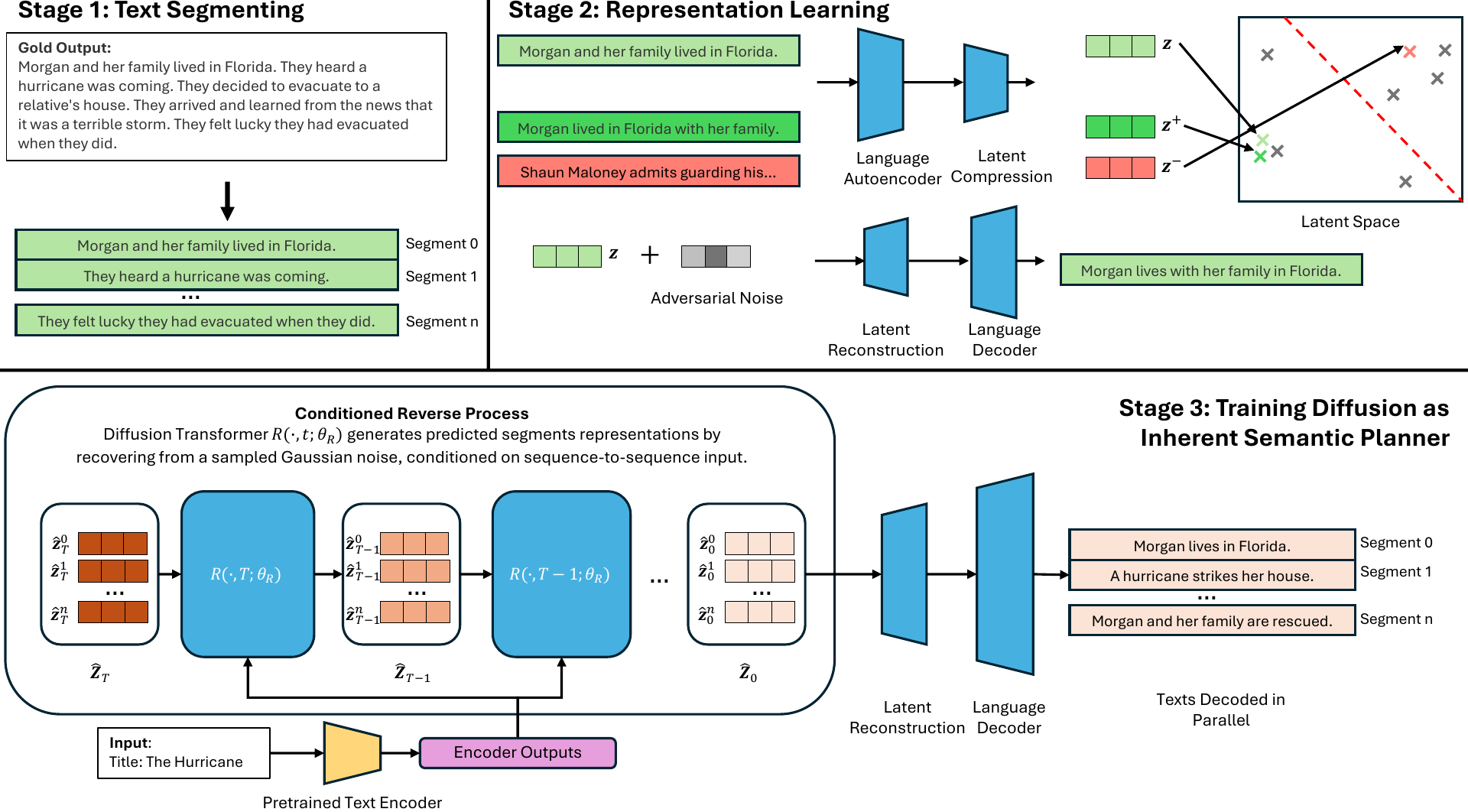}
    \caption{Overview of the training pipeline of SLD. In the first stage, gold output is divided into segments. In the second stage, we use contrastive and adversarial learning to ensure latent representations are robust to drastic semantic changes. Finally, we train a diffusion model as an inherent semantic planner conditioned on given inputs.}
    \label{fig:pipeline}
    \vspace{-3mm}
\end{figure*}

To address the challenges faced by diffusion language models in controllable long-form generation, we propose
Segment-Level Diffusion (SLD). 
In this section, we first provide an overview of the language generation process using diffusion models in latent space, as illustrated in \autoref{fig:model_comparison}. We then introduce our improvements, offering an overview of the three training stages of SLD, as illustrated in \autoref{fig:pipeline}: output segmentation, representation learning, and training diffusion processing for semantic planning. Detailed training algorithm of the our model is outlined in Appendix as Algorithm \ref{alg:sld-training}.

\subsection{Formulation}
Given an input text sequence $\mathbf{i} = \{i^1, i^2, \ldots, i^n\}$ consisting of $n$ tokens and an output sequence $\mathbf{o} = \{o^1, o^2, \ldots, o^m\}$ consisting of $m$ tokens, we model the conditional probability $p(\mathbf{o}|\mathbf{i})$ using a learnable diffusion model $R(;\theta_R)$. We follow \citet{lovelace-etal-2024-diffusion} by introducing additional encoding and decoding components to convert texts into continuous latent representations. The gold output is first encoded into language model hidden states, then projected to latent space as variable $\mathbf{z}$. 
The diffusion model operates on the continuous latent variables $\mathbf{z}$ across $T$ time steps, modelled as a Markov chain \cite{sohl2015deep, ho2020denoising, song2020improved}, and consists of two processes: a backward process for inference and a forward process for training. 

\rparagraph{Inference} The backward process generates the latent representation of the predicted output text $\hat{\mathbf{o}}$ by iteratively removing noise from an initial noisy sample. Starting with a variable $\hat{\mathbf{z}}_T \sim \mathcal{N}(\hat{\mathbf{z}}_T; \mathbf{0}, \mathbf{I})$, the diffusion model with parameters $\theta_R$ predicts the de-noised variable $\hat{\mathbf{z}}_{t-1}$ at each time step $t$ as follows:
\begin{equation}   p(\hat{\mathbf{z}}_{t-1} | \hat{\mathbf{z}}_{t}; {\theta_R}) = \mathcal{N} \left( \hat{\mathbf{z}}_{t-1}; \boldsymbol{\mu}_{\theta_R}^{t-1}, {\boldsymbol{\sigma}_{\theta_R}^{t-1}}^2 \right)
\end{equation}
where $\boldsymbol{\mu}_{\theta_R}$ and $\boldsymbol{\sigma}_{\theta_R}$ are the predicted mean and variance at each time step. The diffusion model, $R(;\theta_R)$, estimates $\hat{\mathbf{z}}_{t-1}$. It conditions on the input sequence $\mathbf{i}$, using the encoder outputs from a pre-trained text encoder $Enc_{ctx}(;\theta_{ctx})$ as below:
\begin{equation}
\hat{\mathbf{z}}_{t-1} =  R(\hat{\mathbf{z}}_t, t, Enc_{ctx}(\mathbf{i};\theta_{ctx}); \theta_{R})
\end{equation}
The model keeps refining the noisy sample $\hat{\mathbf{z}}_T$ with respect to the input sequence $\mathbf{i}$ to recover $\hat{\mathbf{z}}_0$ which will be converted to text. The predicted latent representation $\hat{\mathbf{z}}_0$ is passed to a function parameterised by $\theta_g$, which reconstructs it to match the input dimensions of an AR decoder with parameters $\theta_{dec}$ for decoding:
\begin{equation}
    g(\hat{\mathbf{z}}_0; \theta_g) \in \mathbb{R}^{k \times h_{lm}}, \hat{\mathbf{o}} = Dec(g(\hat{\mathbf{z}}_0; \theta_g); \theta_{dec}).
\end{equation}

\rparagraph{Training} The diffusion model $R(;\theta_R)$ is trained by minimizing a regression loss to predict the noise added during the forward process. In the forward process, an original representation $\mathbf{z}_0$ of encoded $\mathbf{o}$ from a training instance $(\mathbf{i}, \mathbf{o}) \sim \mathcal{D}$ is gradually corrupted into Gaussian noise over $T$ time steps. 
The encoding process consists of an encoder with parameters $\theta_{enc}$ that encodes the output texts:
\begin{equation}
    Enc(\mathbf{o}; \theta_{enc}) \in \mathbb{R}^{m \times h_{lm}}
\end{equation}
and a compression function with parameters $\theta_{f}$ that projects encoder outputs into a length-independent latent space using:
\begin{equation}
    \mathbf{z} = f(Enc(\mathbf{o; \theta_{enc}}); \theta_{f}) \in \mathbb{R}^{k \times h_{rep}}
\end{equation}
Here, we reduce the dimension of encoder outputs to a fixed-length representation with $k \leq m$ and $h_{rep} \ll h_{lm}$. This reduces the difficulty of diffusion model learning from a sparse distribution. The corruption is modelled as:
\begin{equation}
    q(\mathbf{z}_{t} | \mathbf{z}_{t-1}) = \mathcal{N} (\mathbf{z}_{t}; \sqrt{1-\beta_{t}}\mathbf{z}_{t-1}, \beta_{t}\mathbf{I})
\end{equation}
where $\beta_t$ controls the variance of the added noise at each step. The objective is to minimize the distance between the predicted representation $\hat{\mathbf{z}}_t$ and the true posterior $\mathbf{z}_t$ which is computed in closed form by sampling from the forward process. 
\begin{equation}
    \mathcal{L}(\theta_{R}) = \sum_{t=1}^T \underset{q(\mathbf{z}_{t} | \mathbf{z}_0)} {\mathbb{E}} \left\| \hat{\mathbf{z}}_{t} - \mathbf{z}_t \right\|^2_2
\end{equation}
The loss above trains the model to iteratively reverse the corruption applied during the forward process, enabling high-quality data generation in the backward process.

\subsection{Segmented Text Generation}
Inspired by the concept of image patches \cite{ding2023patched}, in the first stage, we segment long outputs into smaller segments, such as sentences or dialogue utterances, rather than projecting the entire output into a single latent space representation. This segmentation effectively reduces the size and complexity of each latent representation, simplifying diffusion predictions and enabling greater flexibility for scaling, allowing the model to handle long-form text more efficiently. Formally, we construct $\mathbf{P} = \{\mathbf{p}^1, \ldots, \mathbf{p}^j\}$,
where each $\mathbf{p}^j$ corresponds to a non-overlapping contiguous
segment of tokens in $\mathbf{o}$. This process yields a set of latent representations $
    \mathbf{Z} = \{\mathbf{z}^1, \ldots, \mathbf{z}^j\}$,
after encoding, establishing a one-to-one correspondence between each segment and its respective latent representation. As shown in \autoref{fig:pipeline}, a story is divided into $n$ segments (sentences in this case), and the diffusion model will predict the same number of representations and decode them into $n$ segments in parallel.

\subsection{Learning Latent Representations for Robust Decoding}

As mentioned earlier, performing diffusion in latent space for text generation requires training the language encoder $Enc(;\theta_{enc})$, latent compression $f(;\theta_f)$, reconstruction $g(;\theta_g)$, and decoder $Dec(;\theta_{dec})$. A straightforward approach is to use the conversion loss incurred during decoding $\hat{\mathbf{p}} = Dec(g(\mathbf{z};\theta_g);\theta_{dec})$, where $\mathbf{z} = f(Enc({\mathbf{p}};\theta_{enc});\theta_{f})$, for a patch of text $\mathbf{p} = \{o_1, \ldots, o_p\}$. We denote the parameters collectively as $\theta_{in} = \{\theta_{enc}, \theta_f\}$ for the encoding and compression, $\theta_{out} = \{ \theta_g, \theta_{dec}\}$ for reconstruction and decoding, and $\theta_{rep} = \theta_{in} \cup \theta_{out}$ for latent representation parameters. The parameters are trained using the standard cross-entropy loss as below.
\begin{equation}
\mathcal{L}_{\text{cnv}}(\theta_{rep}) = - \sum_{l=1}^p \log p(o_l | o_{<l}, \mathbf{p}; \theta_{rep})
\end{equation}

However, learning a robust latent representation is non-trivial. \citet{zhang2024planner} highlighted that a good latent representation should ensure low latent-to-text conversion error and smoothness in the latent distribution. Small perturbations to the latent representation should not significantly affect decoding, while textual segments with similar meanings should be appear relatively close together in the latent space. To achieve this, we augment the cross-entropy loss with contrastive learning and adversarial training as regularization techniques.

\rparagraph{Contrastive Representation Learning}  
In our framework, we operate in latent spaces for segments of text $\mathbf{p}$, which are relatively short compared to the paragraphs in \citet{zhang2024planner}. This allows us to easily obtain meaningful positive examples (\eg paraphrases) $\mathbf{p}^+$ and negative examples (\eg out-of-domain text) $\mathbf{p}^-$ for contrastive learning \cite{gao-etal-2021-simcse}. During training, we derive online representations $\mathbf{z}, \mathbf{z}^+$, and $\mathbf{z}^-$ for these segments and employ the following loss, where $\tau$ is the temperature parameter and $sim()$ is a similarity function (\eg cosine similarity):
\begin{equation}
\mathcal{L}_{\text{cst}}(\theta_{in}) = -\log \frac{e^{(\text{sim}(\mathbf{z}, \mathbf{z}^+)/\tau)}}{e^{(\text{sim}(\mathbf{z}, \mathbf{z}^+)/\tau)} + e^{(\text{sim}(\mathbf{z}, \mathbf{z}^-)/\tau)}}
\end{equation}

\rparagraph{Noise for Decoding Robustness}  
To make the decoding process less sensitive to noise in latent representations, and optimise the decoder with respect to the outputs of the diffusion model instead of the encoding process, we introduce noise into both the input text and latent representations during training. Specifically, we apply the following two noise injection strategies with small probabilities.

First, inspired by \citet{zhang2024planner}, we substitute a small portion of the original input text $\Tilde{\mathbf{p}} = \texttt{sub}(\mathbf{p})$ with uniformly randomly sampled tokens from the model's vocabulary, requiring the  model to exhibit a certain level of error tolerance.

Second, we add adversarial noise to the latent representations $\mathbf{z}$. Extending \citet{miyato2016adversarial}'s adversarial noise approach, we define $\epsilon_{adv}$ as the noise norm. The adversarial noise $\mathbf{r}_{adv}$ is computed as the normalized negative gradient of the loss:
\begin{equation}
\mathbf{r}_{adv} = -\frac{\epsilon_{adv} \mathbf{g}}{||\mathbf{g}||_2} \text{, where }
\end{equation}
\begin{equation}
      \mathbf{g} = \nabla_{\mathbf{p}} \sum_{l=1}^p \log p(o_l | o_{<l}; {\theta}^{\prime}_{out})
\end{equation}
and $p(; {\theta}^{\prime}_{out})$ is modelled with ${\theta}^{\prime}_{out}$ as a frozen copy of the parameters. This approach simulates the ``worst-case'' noise scenario, training the model's reconstruction and decoding network to recover sequences under adversarial conditions with the following loss:
\begin{equation}
\mathcal{L}_{\text{adv}}(\theta_{out}) = - \sum_{l=1}^p \log p(o_l | o_{<l}, \mathbf{z} + \mathbf{r}_{adv}; \theta_{out}).
\end{equation}

\noindent Combining all three losses, the loss function for representation learning is defined as:
\begin{multline}
    \mathcal{L}_{\text{rep}}(\theta_{rep}) = \frac{1}{N}\sum_{\mathbf{p}} ( \mathcal{L}_{\text{cnv}}(\theta_{rep}) \\ + \lambda_1 \mathcal{L}_\text{{cst}}(\theta_{in}) + \lambda_2 \mathcal{L}_\text{{adv}}(\theta_{out}) )
\label{eq:rep_loss}
\end{multline}
where $N$ is the size of the mini-batch, and $\lambda_1, \lambda_2$ are hyperparameters controlling the contribution of contrastive and adversarial losses.

\subsection{Diffusion for Semantic Planning}

After the aforementioned components are trained to convergence, we describe the final stage of training diffusion model as a segment-level semantic planner. The diffusion model leverages learned segment representations to plan and generate meaningful passages consisting of a sequence of segments.  

Given a context $\mathbf{i}$ and collated output texts segments $\mathbf{P}$, we derive the corrupted latent representation of patches $\mathbf{Z}_{t} \in \mathbb{R}^{p \times k \times h_{rep}}$ at time $t$. Absolute positional embedding is applied to the flattened representation with respect to $p \times k$ \cite{vaswani2017attention}. A transformer-based model, which is typically a diffusion transformer (DiT) \cite{peebles2023scalable}, is used for de-noising, defined as:
\begin{equation}
    \hat{\mathbf{Z}}_{t-1} = R(\mathbf{Z}_t, t, Enc_{ctx}(\mathbf{i}; \theta_{ctx}); \theta_R),
\end{equation}
where $Enc_{ctx}(;\theta_{ctx})$ is a pre-trained language encoder with frozen parameters. The encoded outputs serves as the cross-attention target for the diffusion transformer enabling conditional generation. We define the diffusion de-noising loss as:
\begin{equation}
    \mathcal{L}_\text{{noise}}(\theta_R) = \underset{\mathbf{P}, \mathbf{i}, t}{\mathbb{E}} \left\|  \hat{\mathbf{Z}}_{t-1} - \mathbf{Z}_{t-1} \right\|^2_2.
\end{equation}

To strengthen the guidance and ensure  the fluency of the  decoded text, we add a post-diffusion training loss, which incorporates loss signals from the reconstruction and decoding processes. This strategy effectively teaches the diffusion model how to use $g(;\theta_g)$ and $Dec(;\theta_{dec})$, further enhancing the quality of the generated text. Similar to \citet{zhang2024pixel}'s pixel level guidance, we freeze the reconstruction and decoding parameters and define the additional objectives for $\theta_R$ as follows:
\begin{equation}
    \mathcal{L}_{\text{rec}}(\theta_R) = \underset{\mathbf{P}, \mathbf{i}, t}{\mathbb{E}} \left\| g(\hat{\mathbf{Z}}_{t};\theta_g) - g({\mathbf{Z}}_{t};\theta_g) \right\|^2_2,
\end{equation}
\begin{equation}
    \mathcal{L}_{\text{dec}}(\theta_R) = \underset{\mathbf{P}, \mathbf{i}, t}{\mathbb{E}} \left[ - \sum_{l=1}^p \log p(o_l | o_{<l}, \hat{\mathbf{Z}}_t;\theta_{dec}) \right].
\end{equation}

Combining these three losses above with hyperparameters $\lambda_3$ and $\lambda_4$ as weighting factors, we define the diffusion loss function as below.
\begin{equation}
    \mathcal{L}_{\text{diff}}(\theta_R) = \mathcal{L}_\text{{noise}}(\theta_R) + \lambda_3 \mathcal{L}_{\text{rec}}(\theta_R) + \lambda_4 \mathcal{L}_{\text{dec}}(\theta_R)
\label{eq:diff_loss}
\end{equation}

\section{Evaluation}

\subsection{Datasets and Baselines}
We evaluate our implementation on datasets with an increasing number of utterances to assess its performance across various tasks. We start with the XSum dataset \cite{narayan-etal-2018-dont}, consisting of BBC news articles paired with concise, one-sentence summaries, to compare our model against baseline short-form diffusion models. We then scale up to longer outputs using the ROCStories dataset \cite{mostafazadeh-etal-2016-corpus} for title-to-story generation, and the DialogSum dataset \cite{chen-etal-2021-dialogsum} for summary-to-dialogue generation. These datasets allow us to evaluate the model's capability for long-form generation. Additionally, we test our model on dialogue  generation with DeliData \cite{karadzhov2023delidata}. This dataset consists of multi-party problem-solving dialogues during which participants propose solutions that are scored for their correctness. By comparing the predicted user score trajectories against the ground truth and identifying hallucinations, we analyse the effectiveness of applying control to the model's generation. 

We compare our model against a range of baselines. Specifically, we use LD4LG \cite{lovelace2024latent} as the diffusion baseline and Flan-T5 Large \cite{chung2024scaling} as the autoregressive baseline. For the XSum dataset, we also compare against the token-level diffusion model GENIE \cite{lin2023text}. For the DialogSum dataset, we include comparisons with the dialogue-level diffusion model Diffuse-CG \cite{chen2023controllable}.

\subsection{Evaluation Metrics}
We use ROUGE as the primary evaluation metric to assess the quality and similarity of generated text with respect to the gold output. While ROUGE provides a baseline for lexical overlap, we acknowledge its limitations in capturing semantic fidelity, coherence, and conversational nuances, particularly in controlled long-form generation.  To address this, we extend human evaluation guidelines from \citet{clark-etal-2023-seahorse}, assessing repetition, fluency, coherence, compatibility (ROCStories/DialogSum), and hallucination (DeliData):
\begin{itemize}[noitemsep, topsep=0pt] \item \textbf{Repetition}: Check for repetitive tokens or utterances that affect meaning. \item \textbf{Fluency/Grammar}: Assess grammatical correctness and fluency. \item \textbf{Coherence}: Evaluate logical flow and naturalness of interactions. \item \textbf{Compatibility}: Ensure alignment with the story title/dialogue summary. \item \textbf{Hallucination}: Detect impossible choices or non-existent participants. \end{itemize}

Human scores range from 0 to 3, with higher scores indicating better performance. Details and examples are in Appendix \ref{appendix:human_eval}. Following the evaluation metrics in the literature, we also evaluate the perplexity of generated text using GPT-2 Large \cite{radford2019language} as teacher model, and record the average length of generated texts.

\subsection{Implementation Details}

We build upon the design of Latent Diffusion for Language Generation (LD4LG) proposed by \citet{lovelace2024latent}, using Flan-T5 Base \cite{chung2024scaling} as the backbone to initialize our encoder and decoder. We incorporate the Perceiver Resampler \cite{alayrac2022flamingo} as the compression and reconstruction unit and employ a pre-LayerNorm transformer as the de-noising model \cite{vaswani2017attention}. For contrastive learning targets, we use Llama-3-8B-Instruct \cite{grattafiori2024llama3herdmodels} to generate paraphrases for each text segment. For XSum, we sampled out-of-domain (OOD) texts from Movie-Dic \cite{banchs-2012-movie} dataset as hard negative targets. For other datasets, we sample from CNN/Daily Mail \cite{see-etal-2017-get}.


\section{Results}
\begin{table*}[h!]
\centering
\small

\textbf{XSum} \\[1mm]
\begin{tabular}{p{0.18\textwidth}|p{0.1\textwidth}|p{0.1\textwidth}|p{0.1\textwidth}|p{0.1\textwidth}|p{0.1\textwidth}|p{0.1\textwidth}}
\toprule
\textbf{Models}  & \textbf{ROUGE-1} & \textbf{ROUGE-2} & \textbf{ROUGE-L} & \textbf{Length} & \textbf{PPL} & \textbf{Fluency$^*$} \\ 
\midrule
Gold             & N/A              & N/A              & N/A              & 21.25           & 22.93          & 2.98   \\ \midrule
Flan-T5    & \textbf{44.31}            & \textbf{20.44}            & \textbf{35.76}            & 19.25           & \textbf{18.02}          & \textbf{2.85}   \\ 
GENIE            & 36.14            & 12.63            & 28.37            & 20.39           & 145.37         & 1.78   \\ LD4LG            & 32.96            & 11.70            & 26.74            & 20.43           & 44.33          & 2.83   \\ 
Ours & 34.24            & 12.38            & 27.77            & 21.27           & 46.09          & 2.78   \\ 
\bottomrule
\end{tabular}
\vspace{1mm}

\textbf{ROCStories} \\[1mm]
\begin{tabular}{p{0.11\textwidth}|p{0.1\textwidth}|p{0.07\textwidth}|p{0.06\textwidth}|p{0.1\textwidth}|p{0.1\textwidth}|p{0.1\textwidth}|p{0.12\textwidth}}
\toprule
\textbf{Models} & \textbf{ROUGE-L} & \textbf{Length} & \textbf{PPL} & \textbf{Repetition (mean, var)} & \textbf{Fluency (mean, var)} & \textbf{Coherence (mean, var)} & \textbf{Compatibility (mean, var)} \\ 
\midrule
Gold             & N/A             & 42.53           & 20.99        & 2.93, 0.08                      & 2.87, 0.16                   & 2.86, 0.17                      & 2.74, 0.34                       \\ \midrule
Flan-T5   & 16.46           & 32.40           & \textbf{9.14}         & 2.41, 0.82                      & \textbf{2.62, 0.59}                   & \textbf{2.33, 0.78}                      & 2.07, 0.83                       \\ 
LD4LG            & \textbf{16.57}           & 36.56           & 65.32        & 2.14, 1.26                      & 1.78, 1.02                   & 1.54, 0.92                      & 1.79, 1.17                       \\ 
Ours              & 16.13           & 40.70           & 43.67        & \textbf{2.45, 0.89}                      & 2.41, 0.83                   & 2.10, 0.88                      & \textbf{2.42, 0.88}                       \\ 
\bottomrule
\end{tabular}

\vspace{1mm}

\textbf{DialogSum} \\[1mm]
\begin{tabular}{p{0.12\textwidth}|p{0.1\textwidth}|p{0.06\textwidth}|p{0.06\textwidth}|p{0.1\textwidth}|p{0.1\textwidth}|p{0.1\textwidth}|p{0.12\textwidth}}
\toprule
\textbf{Models} & \textbf{ROUGE-L} & \textbf{Length} & \textbf{PPL} & \textbf{Repetition (mean, var)} & \textbf{Fluency (mean, var)} & \textbf{Coherence (mean, var)} & \textbf{Compatibility (mean, var)} \\ 
\midrule
Gold                        & N/A      & 117.32  & 9.68  & 2.95, 0.07 & 2.86, 0.14 & 2.91, 0.11 & 2.92, 0.11 \\ \midrule
Flan-T5               & 26.34    & 131.92 & \textbf{3.78}  & 1.83, 0.82 & 2.60, 0.55 & 2.07, 0.64 & 2.27, 0.60 \\ 
LD4LG                       & 20.90    & 81.28   & 43.82 & 1.93, 1.32 & 1.43, 0.84 & 1.39, 0.90 & 1.61, 1.16 \\ 
Ours          & \textbf{27.97}    & 113.92  & 16.39 & \textbf{2.80, 0.23} & \textbf{2.83, 0.17} & \textbf{2.40, 0.45} & \textbf{2.57, 0.37} \\ 
Diffuse-CG$^{**}$         & 27.57    & 84.23   & 68.45 & N/A         & N/A         & N/A         & N/A         \\
\bottomrule
\end{tabular}

\vspace{1mm}

\textbf{DeliData} \\[1mm]

\begin{tabular}{p{0.11\textwidth}|p{0.1\textwidth}|p{0.07\textwidth}|p{0.06\textwidth}|p{0.1\textwidth}|p{0.1\textwidth}|p{0.1\textwidth}|p{0.12\textwidth}}
\toprule
\textbf{Models} & \textbf{ROUGE-L} & \textbf{Length} & \textbf{PPL} & \textbf{Repetition (mean, var)} & \textbf{Fluency (mean, var)} & \textbf{Coherence (mean, var)} & \textbf{Hallucination (mean, var)} \\ 
\midrule
Gold             & N/A              & 53.04           & 13.42        & 2.60, 0.46                      & 2.66, 0.42                   & 2.55, 0.47                      & 2.85, 0.29               \\ \midrule
Flan-T5    & 25.83            & 234.35          & \textbf{9.79}         & 2.27, 0.50                      & 2.59, 0.39                   & \textbf{2.30, 0.43}                      & \textbf{2.48, 0.52}                       \\
LD4LG            & 21.14            & 68.08           & 51.10        & \textbf{2.52, 0.89}             & 2.06, 0.80                   & 1.78, 0.71                      & 1.63, 0.87                       \\ 
Ours              & \textbf{30.51}   & 71.61           & 13.41        & 2.50, 0.54                      & \textbf{2.60, 0.34}          & \textbf{2.30, 0.47}            & \textbf{2.48, 0.55}                       \\ 
\bottomrule
\end{tabular}

\caption{Comparison of results on the XSum, ROCStories, DialogSum, and DeliData datasets. Fluency$^*$ for XSum is calculated using Gemini-1.5 \cite{team2024gemini} as LLM-as-a-Judge with details are provided in Appendix \ref{appendix:human_eval}. Results for Diffusion-CG$^{**}$ are directly taken from \citet{chen2023controllable}.}

\label{tab:combined_tables}
\vspace{-3mm}
\end{table*}

We present the results of our model against the baselines on XSum, ROCStories, DialogSum and DeliData in \autoref{tab:combined_tables}. For short-form generation task, XSum ($\sim$ 30 tokens), our model demonstrates on-par performance compared with other baselines. For long-form generation ($\geq$ 50 tokens) with naturally occurring segmentation (sentences in ROCStories, utterances in DialogSum and DeliData), our model shows better overall performance, especially in repetition, fluency and compatibility.

\rparagraph{Short-form} Although Flan-T5 achieves the highest ROUGE scores on XSum, SLD achieves on-par performance compared to other diffusion methods while maintaining a length closer to the reference. Importantly, the results indicate the importance of an AR decoder. For Flan-T5, LD4LG and SLD that models $p(o_l|o_{<l},\mathbf{i})$, they have substantially higher fluency scores than GENIE which uses diffusion models to decode token level embeddings directly, modelling $p(\mathbf{o}|\mathbf{i})$. In addition, autoregressive models can terminate generation early, while GENIE always has to predict up to its 64 token with unnecessary paddings due to its NAR nature.

\rparagraph{Long-form} Our method consistently achieves the highest ROUGE-L scores for summary-to-dialogue and multi-party dialogue generation tasks, outperforming all baselines in datasets such as DialogSum and DeliData. In terms of human evaluation metrics, SLD shows clear advantages in reducing repetition and enhancing compatibility, producing logically consistent and contextually accurate outputs across tasks like ROCStories and DialogSum. This demonstrates diffusion guidance for controlled generation makes our outputs adhere to the given topic better. Furthermore, SLD maintains competitive fluency and coherence, matching autoregressive models like Flan-T5, while LD4LG's performance drops as output sequence gets longer. SLD also demonstrates robustness in handling complex dialogue structures, aligning well with user score trajectory distributions in multi-party settings, as detailed in Appendix \ref{appendix:delidata_dist}. Additionally, SLD's outputs closely match the desired length and structure of gold references, showcasing its adaptability across diverse text generation tasks. 

Overall, SLD improves across datasets and metrics, particularly in long-form, contextually accurate text generation. While Flan-T5 excels in metrics like perplexity, it suffers from catastrophic repetition after fine-tuning. LD4LG struggles with coherence and fluency in longer texts due to abrupt semantic shifts in latent representations, as reflected by high variance in human evaluations. In contrast, SLD achieves better human scores, particularly in fluency, coherence, and compatibility.

\section{Analysis}

\rparagraph{Representation to Text}
To assess representation learning, we tested various autoencoder-decoder configurations using BLEU \cite{papineni-etal-2002-bleu} to compare input text with recovered text after encoding. Using DialogSum utterances as segments, we evaluated LD4LG, ML-Planner, and SLD with and without contrastive learning. For consistency, segments were limited to 64 tokens, represented in a latent space of $32 \times 64$. An LD4LG baseline with longer dialogues (up to 512 tokens) and latent dimensions of $256 \times 64$ was also included.

\begin{figure}
    \centering
    \includegraphics[width=0.45\textwidth]{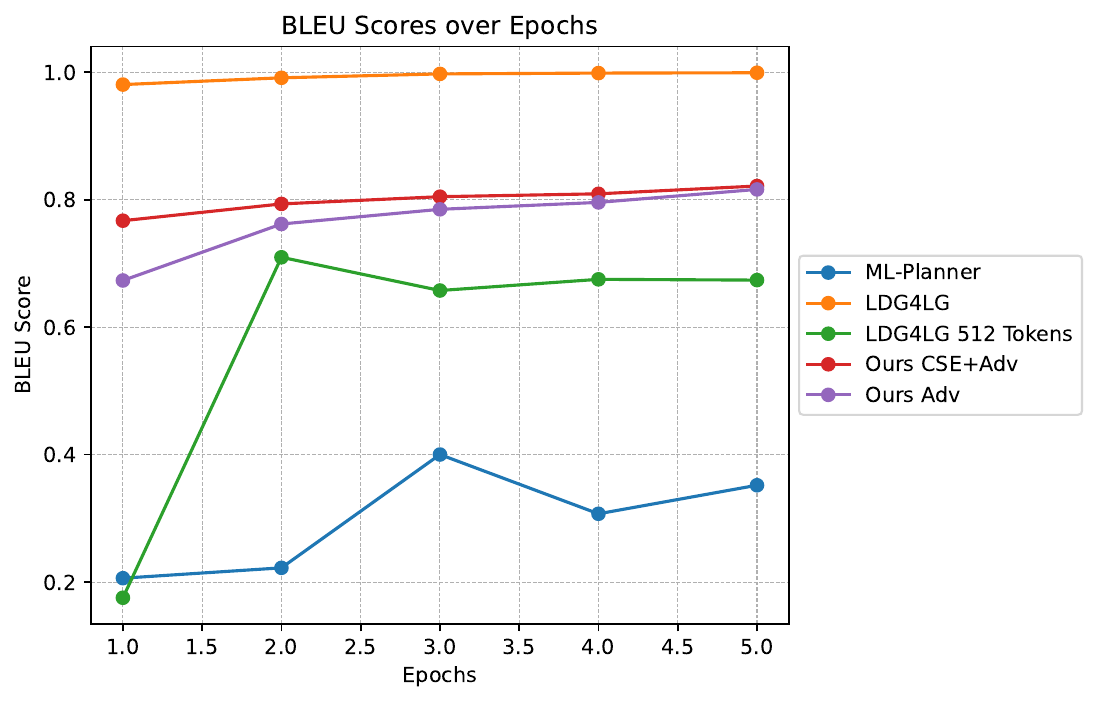}
    \caption{BLEU score of different auto-encoder/decoder models for text conversion on DialogSum dataset of a single utterance.}
    \label{fig:bleu}
\end{figure}

\autoref{fig:bleu} shows that ML-Planner failed to converge within five epochs, while LD4LG achieved a BLEU score of $1.00$, indicating perfect recovery. However, LD4LG’s performance degraded when scaled to longer texts, highlighting limitations in generalization. Without contrastive learning, our model occasionally corrupted words, altering meaning. Incorporating contrastive learning enabled meaningful paraphrases instead of semantic corruption, as demonstrated in \autoref{tab:text_quality_comparison}, emphasizing its role in enhancing representation quality.

\rparagraph{Decoding after De-noising}
We further investigated how latent representations behave under perturbations to evaluate their robustness during the de-noising process. We randomly selected 100 sentences from the ROC dataset, along with their paraphrases and OOD sentences sampled from the CNN/Daily Mail dataset. We visualized a 2D PCA projection of learned representations, detailed in \autoref{appendix:albation}. Without contrastive learning, the representations of original sentences and OOD sentences showed significant overlap, increasing the risk of abrupt semantic changes during decoding. In contrast, representations learned with contrastive training were better clustered and distinct, providing improved robustness for diffusion predictions.

To investigate further, we sampled the de-noising trajectory of a test sentence and analysed the decoded text along the trajectory. Representations trained with both contrastive learning and adversarial training produced text that was more robust to noise and less prone to abrupt semantic shifts. This robustness facilitates smoother and more reliable predictions during the diffusion process.

\begin{table}
\centering
\footnotesize
\begin{tabular}{p{0.17\columnwidth}|p{0.7\columnwidth}}
\toprule
\textbf{Model} & \textbf{Generated Text} \\ \midrule
\texttt{Gold} & \#Person1\#: What made you decide on this type of occupation? \\ \midrule
\texttt{Planner} & \#Person1\#: \textcolor{Red}{I'm afraid I can't.} \\ \midrule
\texttt{LD4LG} & \#Person1\#: What made you decide on this type of occupation? \\ \midrule
\texttt{Adv} & \#Person1\#: What's \textcolor{Orange}{your} decide on this type of occupation? \\ \midrule
\texttt{Adv+CSE} & \#Person1\#: What made you decide on this type of \textcolor{ForestGreen}{job}? \\ \midrule
\end{tabular}
\caption{Text quality comparison of different models for text generation. Red indicates wrong conversion, orange indicates corrupted token and green indicates admissible paraphrase.}
\vspace{-3mm}
\label{tab:text_quality_comparison}
\end{table}

\rparagraph{Knowledge Preservation} We additionally evaluate how representation learning and diffusion training affect the knowledge and reasoning abilities of the language model backbone. We test our model against baselines on Flan-T5's pre-training task, ECQA \cite{aggarwal2021explanations}, a knowledge QA dataset with Chain-of-Thought (CoT) reasoning. We measure model performance by prediction accuracy. We ensure fair comparisons by using the same one-shot CoT exemplar setup for Flan-T5 Base and Large \cite{chung2024scaling}. We train LD4LG and our SLD model with the same hyperparameters as ROCStories.

\begin{table}[h]
\small
    \centering
    \begin{tabular}{p{0.12\textwidth}|p{0.07\textwidth}|p{0.07\textwidth}|p{0.10\textwidth}}
    \toprule
    \textbf{Model} & \textbf{Acc} & \textbf{MAUVE} & \textbf{ROUGE-L} \\ 
    \midrule
    Flan-T5 Base  & 0.75 & 0.13 & 33.88  \\
    Flan-T5 Large & \textbf{0.88} & 0.21 & 35.86  \\
    LD4LG         & 0.63 & 0.74 & 38.66  \\
    Ours          & 0.723 & \textbf{0.75} & \textbf{40.14}  \\
    \bottomrule
    \end{tabular}
    \caption{Performance comparison across models on ECQA.}
    \label{tab:model_performance}
    \vspace{-3mm}
\end{table}

Table \ref{tab:model_performance} indicate that SLD preserves the knowledge in its Flan-T5 Base backbone. The diffusion model learns to predict latent representations of unseen concepts within the backbone's original knowledge almost perfectly. Due to the backbone's limitations, all models fail to answer questions about unknown concepts, as shown in Appendix \autoref{tab:ecqa_all_wrong}. This finding provides insights for using existing pre-trained LLM to support diffusion language modelling efficiently. LD4LG frequently struggles to select the correct answer from the provided choices, as illustrated in Appendix \autoref{tab:ecqa_ld4lg_wrong}, highlighting the importance of our text segmentation. While both LD4LG and SLD achieve high MAUVE and ROUGE-L scores due to direct training on the dataset, SLD outperforms LD4LG, indicating the advantage of our post-diffusion control.
\section{Conclusion}
We propose Segment-Level Diffusion (SLD) for controllable long-form text generation using latent space diffusion. Key innovations include text segmentation, robust representation learning via adversarial and contrastive training, and improved latent-space guidance. SLD addresses challenges in generating fluent, coherent, and contextually accurate long-form text, bridging the gap between latent diffusion models and practical long-form generation. It offers a scalable framework for applications like story and dialogue generation. Our results highlight SLD's potential as a new paradigm for controllable text generation and provide insights for future diffusion-based language models.

\section*{Limitations}

This work focuses exclusively on text generation in English, leaving the model's potential for multilingual tasks unexplored. Furthermore, our experiments and evaluations did not involve real-world use cases, limiting insights into practical applicability. Future research could extend our approach to multilingual and application-oriented scenarios such as outline-controlled generation \cite{li-etal-2024-advancing, lee2024navigating}. Additionally, we did not explicitly examine the relationship between the reduced dimensionality of the length-independent latent representations and the original dimensionality of encoded text segments of varying lengths. Due to limited computational capacity, we did not run hyperparameter search on loss ratios but chosen them empirically based on model's performance on the development set. These hyperparameters were chosen empirically, without a systematic exploration of their impact. Future work could leverage principles from information theory \cite{tishby2015deep} to quantify the information capacity of these representations and to balance compression and utility more effectively. Developing a generalized framework to streamline hyperparameter selection across diverse datasets and pre-training tasks would also enhance the scalability of our method. Finally, our modular training approach, where individual components are optimized separately, may introduce suboptimal performance during inference due to error propagation and misalignment between training and inference objectives. Future work could explore end-to-end training strategies to jointly optimize all components, reducing such discrepancies and improving overall performance. Due to inconsistent sampling strategies and output windows of token-level diffusion and other baseline models, we did not perform inference speed comparison.

\section*{Ethics Statement}

No personally identifiable information (PII) was collected or processed during the human evaluation, and all data handling adhered to the General Data Protection Regulation (GDPR) and the University's research guidelines\footnote{\url{https://www.research-integrity.admin.cam.ac.uk/academic-research-involving-personal-data}}. Participants were recruited from within the University on a voluntary basis and were assigned anonymized random IDs to ensure their privacy during the evaluation process. Additionally, all data points presented to annotators were manually reviewed to ensure fairness and accuracy in assessing our methods and to minimize potential bias or harm to participants. This approach reflects our commitment to ethical research practices and to safeguarding the well-being and integrity of participants throughout the evaluation process.

\section*{Acknowledgements}

Georgi Karadzhov was funded by an unrestricted gift from Google on multiparty dialogue agents. 
Andreas Vlachos and Chenxi Whitehouse were supported by the ERC grant AVeriTeC (GA 865958).

\bibliography{anthology,custom}
\bibliographystyle{acl_natbib}

\section*{Appendix}
\appendix

\section{Training} All experiments were conducted on a single NVIDIA A100-SXM-80GB GPU. Training for latent representation learning took approximately $8$–$10$ hours per dataset, while diffusion training required $36$–$60$ hours per dataset. Including inference and other experiments, the total GPU usage amounted to around $500$ hours. The detailed training algorithm is presented in Algorithm \ref{alg:sld-training}. Our code is available at \url{https://github.com/SpaceHunterInf/Segment_Level_Diffusion}.

\rparagraph{Loss Ratio} In representation learning phase, contrastive loss with hyperparameter $\lambda_1$ in \autoref{eq:rep_loss} helps distinguish between nuanced representations, which is crucial in domains with subtle differences in meaning between text segments. For domains like DeliData, where dialogue actions are highly structured and limited (e.g., proposing choices, picking cards, discussing, revising, and submitting), contrastive loss can be reduced or even set to zero (e.g., $\lambda_1=0.1 \text{ or } 0$). Adversarial loss in this same equation controls the model's tolerance to noise, ensuring robust latent representations even in challenging cases. A large $\lambda_2$ can hinder the learning of meaningful representations, but our experiments suggest that moderate values (e.g., $\lambda_2 \approx 0.2$) are crucial for domains with longer text segments, such as DialogSum, where segments (e.g., multi-sentence utterances) require higher noise tolerance to maintain generation quality. Tasks with shorter segments or clearer structure might allow for lower $\lambda_2$ values. In \autoref{eq:diff_loss} for diffusion training, $\lambda_3$ controls the loss ratio of reconstruction back to the language model’s hidden state dimensions and $\lambda_4$ corresponds to the reconstruction back to the original text. We provide the hyperparameters we've chosen for each dataset in the table below.

\begin{table}[h]
\small
    \centering
\begin{tabular}{p{0.11\textwidth}|p{0.05\textwidth}|p{0.05\textwidth}|p{0.05\textwidth}|p{0.05\textwidth}}
\toprule
\textbf{Dataset} & $\lambda_1$ & $\lambda_2$ & $\lambda_3$ & $\lambda_4$ \\ 
\midrule
XSum        & 0.5  & 0.2  & 0.5  & 0.5\\ 
ROCStories  & 0.2  & 0.2  & 0.5  & 0.5\\ 
DialogSum   & 0.5  & 0.2  & 0.5  & 0.5\\ 
Delibot     & 0    & 0.2  & 0.5  & 0.5\\ 
\bottomrule
\end{tabular}

\caption{Hyperparameter choices of loss ratio.}
\vspace{-3mm}
\label{tab:hyperparameter}
\end{table}

\rparagraph{Inference Sampling} We use the sampling algorithm as \citet{lovelace2024latent}, a DDPM pytorch implementation\footnote{\url{https://github.com/lucidrains/denoising-diffusion-pytorch}}. For training we use $2000$ steps in forward process as target with cosine schedule. For inference, we use spaced diffusion with only $250$ steps.

\section{Ablation Study: Latent Representation} \label{appendix:albation} 
\begin{figure*}
    \centering
    \includegraphics[width=1.0 \textwidth]{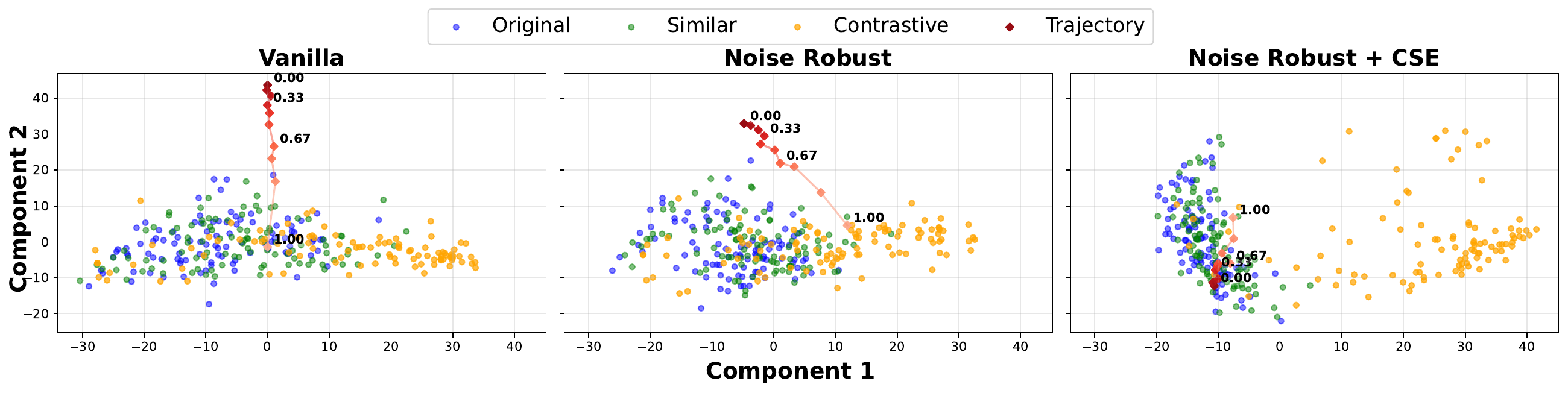}
    \caption{Comparison of PCA 2D projections of latent representations for sampled segmented sentences from ROCStories (Blue), their paraphrases (Green), and out-of-domain (OOD) sentences sampled from CNN/Daily Mail (Orange) under three training paradigms: Vanilla training, Noise Robust training, and Noise Robust + Contrastive learning. The red trajectory illustrates the denoising path of the sentence 'David noticed he had put on a lot of weight recently.' The trajectory is annotated with noise ratios, where 1.0 (Lighter Red) represents pure Gaussian noise and 0.0 (Darker Red) indicates no noise. }
    \label{fig:rep_comp}
    \vspace{-5mm}
\end{figure*}

We conducted an ablation study to evaluate how different representation learning methods affect the noised latent representations during the diffusion process. The "Vanilla" configuration corresponds to the original LD4LG implementation. As shown in Figure \ref{fig:rep_comp} and Table \ref{tab:decoded_text}, our method achieves a smoother latent distribution and more robust representations, resulting in improved diffusion predictions and decoding.

\section{Human Evaluation} \label{appendix:human_eval} For ROCStories, DialogSum, and DeliData, we recruited $5$ participants to evaluate $25$ dialogues generated by LD4LG, Flan-T5, and SLD for each dataset. Annotators also rated the gold dialogue output as a reference. In total, we collected $3 \times 25 \times 4 \times 5 = 1,500$ data points across $4$ evaluation criteria. Before presenting the dialogues, annotators were provided with instructions (example shown in Table \ref{tab:instructions}).

\begin{figure*}
    \centering
    \includegraphics[width=1.0 \textwidth]{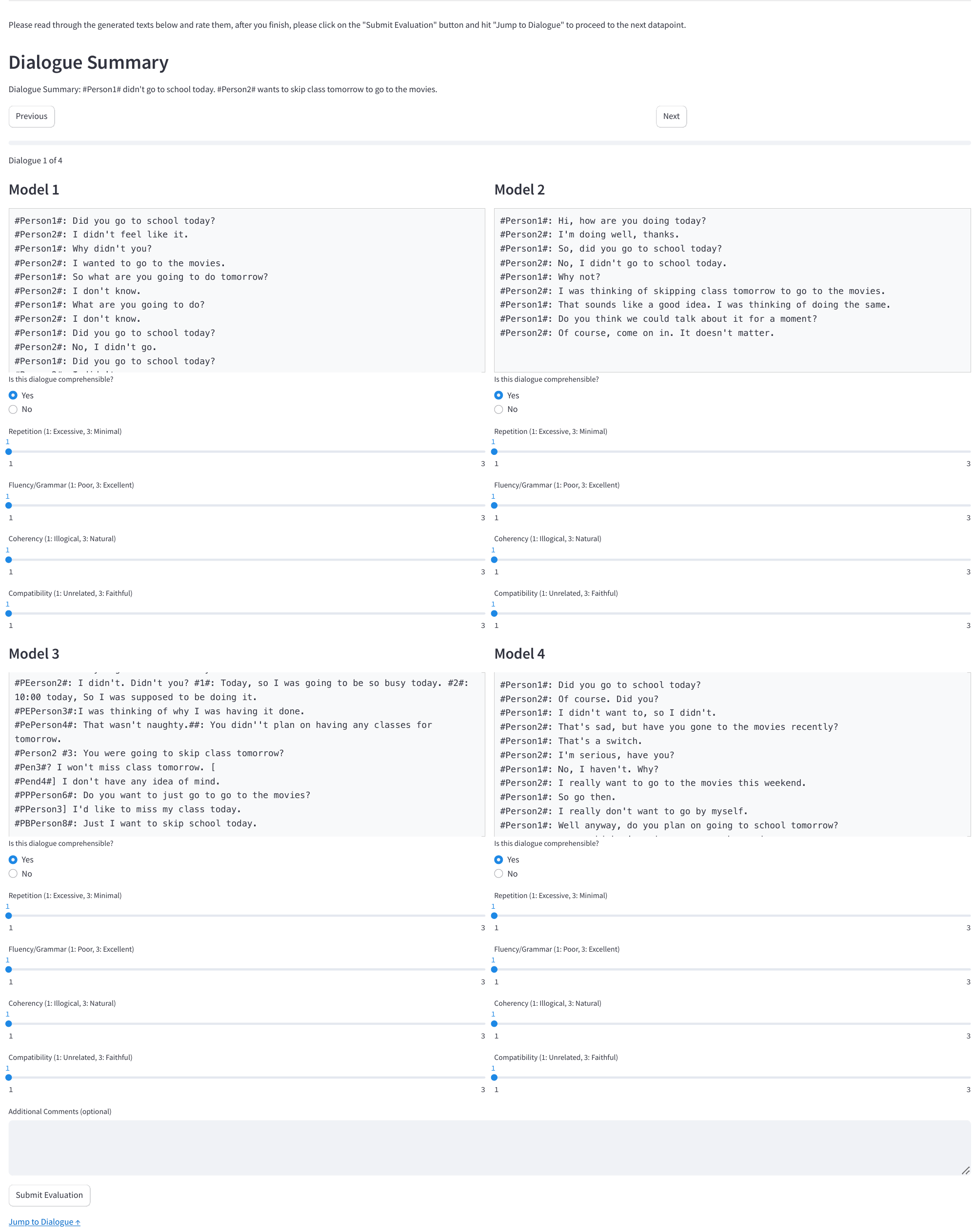}
    \caption{An example of human evaluation interface of DialogSum. Each session starts with a comprehensive instructions with examples, followed by model outputs and questions.}
    \label{fig:web_interface}
\end{figure*}

The evaluation process begins by assessing readability. If a dialogue is deemed unreadable, all criteria are automatically scored as $0$. Otherwise, the dialogue is rated with a minimum score of $1$. An example of the web interface is provided in Figure \ref{fig:web_interface}.

Additionally, we employed Gemini-1.5 \cite{team2024gemini} as an LLM-based evaluator. Using a carefully designed prompt (Table \ref{tab:llm_judge}), Gemini-1.5 provided fluency ratings for predictions from the XSum dataset, achieving high alignment with human annotators.

\section{Case Study: DeliData} \label{appendix:delidata_dist}

\begin{figure}[!htp]
    \centering
    \includegraphics[width=0.4\textwidth]{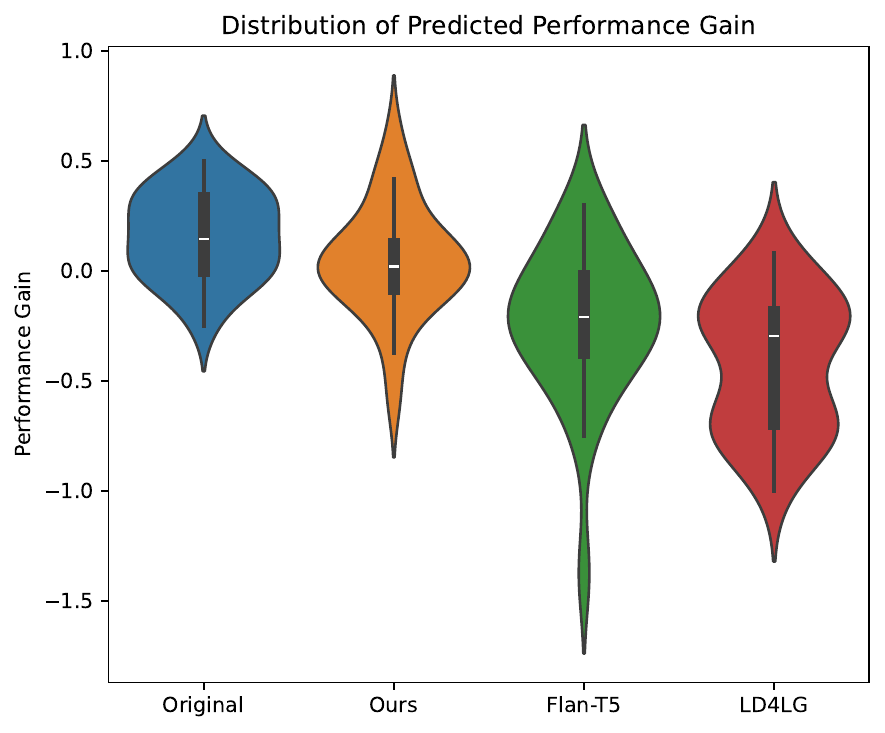}
    \caption{Violin plot of performance gain distribution of the DeliData dialogue continuations. Our model demonstrate a closer distribution with respect to the gold distribution, demonstrating the output is better controlled.}
    \label{fig:delidata_dist}
    \vspace{-5mm}
\end{figure}

To evaluate the controlled generation capabilities of our model, we performed a fine-grained score trajectory analysis on DeliData. This involved comparing the distribution of performance gains at the end of group discussion dialogues, based on users' choices before and after revisions. Following the guidelines of \citet{karadzhov2023delidata}, we observed that dialogue continuations generated by our model produced a performance gain distribution closer to the ground truth, as shown in Figure \ref{fig:delidata_dist}.

\section{Sampled Generations} We provide sample output comparisons in Tables \ref{tab:roc_comparison}, \ref{tab:dialsum_comparison}, and \ref{tab:delidata_comparison}.

\begin{table*}
    \centering
    \footnotesize
    \begin{tcolorbox}[colback=white, colframe=black, title = {Example of Human Evaluation Instructions}]
\textbf{5. Compatibility}\\
- Evaluate whether the dialogue accurately reflects the facts provided in the summary. Rate from 1 to 3:\\
  - \textbf{1}: The dialogue significantly deviates from the summary, either by altering key facts or diverging from the main topic.\\
  - \textbf{3}: The dialogue faithfully reflects the summary, covering all key ideas accurately.\\

\textbf{Examples:}\\
- \textbf{Summary:} Anna asks Bob about his vacation, and Bob talks about visiting a tropical island.\\
  - \textbf{Rating 1:}\\
    - \#Anna\#: "How was your weekend?"\\
    - \#Bob\#: "I stayed home and watched TV."\\
    - \textbf{Rating:} 1 (Completely unrelated to the summary).\\

  - \textbf{Rating 2:}\\
    - \#Anna\#: "How was your vacation?"\\
    - \#Bob\#: "It was great. I visited some nice places."\\
    - \textbf{Rating:} 2 (Partially reflects the summary but lacks details about the tropical island).\\

  - \textbf{Rating 3:}\\
    - \#Anna\#: "How was your vacation?"\\
    - \#Bob\#: "It was amazing! I visited a beautiful tropical island and spent my days snorkeling and relaxing on the beach."\\
    - \textbf{Rating:} 3 (Accurately reflects the summary and includes key details).
\end{tcolorbox}
\caption{An example of human evaluation instructions for evaluating compatibility for DialogSum.}
\label{tab:instructions}
\end{table*}

\begin{table*}
\footnotesize
    \centering
    \begin{tcolorbox}[colback=white, colframe=black, title = {LLM-as-a-judge Prompt}]

\begin{itemize}[leftmargin=*]
    \item \textbf{Rate the grammatical quality of the summary on a scale of 1 to 3:}
    \begin{itemize}[leftmargin=*]
        \item \textbf{1}: Excessive grammatical and clerical errors, making the summary unnatural.
        \item \textbf{3}: Completely fluent with no grammatical errors.
    \end{itemize}
\end{itemize}

\textbf{Examples}

\textbf{Rating 1:}
\begin{itemize}[leftmargin=*]
    \item \textit{Summary:} Blaenau hub to is set up to tech of finger printer securety.
    \item \textbf{Rating}: $1$ (Frequent grammatical errors make the summary unnatural).
\end{itemize}

\textbf{Rating 2:}
\begin{itemize}[leftmargin=*]
    \item \textit{Summary:} A hub for fingerprints tech secure property will be setted in Blaenau Gwent.
    \item \textbf{Rating}: $2$ (Some grammatical errors, but it’s understandable).
\end{itemize}

\textbf{Rating 3:}
\begin{itemize}[leftmargin=*]
    \item \textit{Summary:} A hub for developing fingerprint security technology is being set up in Blaenau Gwent.
    \item \textbf{Rating}: $3$ (Completely fluent and grammatically correct).
\end{itemize}

\textbf{Instructions:}

You are a skilled text quality evaluator specializing in assessing fluency. \\
Your task is to evaluate the fluency of the following text on a scale from 1 to 3 with respect to the examples above:

\begin{quote}
\texttt{\{text\}}
\end{quote}

\begin{itemize}[leftmargin=*]
    \item 1: Poor fluency (e.g., disjointed, unclear, or grammatically incorrect).
    \item 2: Moderate fluency (e.g., some grammatical errors or awkward phrasing but mostly understandable).
    \item 3: High fluency (e.g., clear, smooth, and grammatically correct).
\end{itemize}
Only return the numerical score, enclosed by dollar signs (\$\$), without any additional commentary or explanation.

    \end{tcolorbox}
    \caption{Prompt for Gemini-1.5 Pro to evaluate XSum fluency.}
    \label{tab:llm_judge}
\end{table*}

\begin{algorithm*} 
\small
\caption{Training Algorithm for Segment-Level Diffusion (SLD)}
\label{alg:sld-training}
\SetAlgoLined
\KwIn{
    Input-output text pairs $(\mathbf{i}, \mathbf{o})$, frozen context encoder with parameter $\theta_{ctx}$,\\
    model parameters $\theta_{in} = \{\theta_{enc}, \theta_f\}$ for the encoding and compression, \\$\theta_{out} = \{ \theta_g, \theta_{dec}\}$ for reconstruction and decoding, where $\theta_{enc}, \theta_{dec}$ are initialised from pre-trained language encoder decoder, and $\theta_{rep} = \theta_{in} \cup \theta_{out}$,\\
    hyperparameters $\lambda_1$, $\lambda_2$, $\lambda_3$, $\lambda_4$, 
    temperature $\tau$, 
    batch size $N$, 
    maximum diffusion steps $T$.
}
\KwOut{Trained latent space diffusion model.}
\BlankLine
\textbf{Training: Latent Representation Learning} \\
\For{each mini-batch of text pairs $(\mathbf{i}, \mathbf{o})$}{
    \textbf{1. Output Utterance Segmentation:} Segment $\mathbf{o}$ into patches $\mathbf{P} = \{\mathbf{p}^1, \ldots, \mathbf{p}^j\}$.\\
    \textbf{2. Latent Representation Encoding:} Encode patches into latent space $\mathbf{Z} = \{\mathbf{z}^1, \ldots, \mathbf{z}^j\}$.\\
    \textbf{3. Compute Conversion Loss:} 
    \[
    \mathcal{L}_{\text{cnv}}(\theta_{rep}) = - \sum_{l=1}^p \log p(o_l | o_{<l}, \mathbf{p}; \theta_{rep})
    \]
    \textbf{4. Contrastive Representation Learning:} 
    Sample positive and negative examples $(\mathbf{p}^+, \mathbf{p}^-)$ and compute: 
    \[
    \mathcal{L}_{\text{cst}}(\theta_{in}) = -\log \frac{e^{(\text{sim}(\mathbf{z}, \mathbf{z}^+)/\tau)}}{e^{(\text{sim}(\mathbf{z}, \mathbf{z}^+)/\tau)} + e^{(\text{sim}(\mathbf{z}, \mathbf{z}^-)/\tau)}}
    \]
    \textbf{5. Noise Robustness Training:} 
    Add random noise to patches and latent space $\mathbf{z}$:
    \[
    \mathcal{L}_{\text{adv}}(\theta_{out}) = - \sum_{l=1}^p \log p(o_l | o_{<l}, \mathbf{z} + \mathbf{r}_{adv}; \theta_{out}),
    \]
    where $\mathbf{r}_{adv}$ is adversarial noise computed using a frozen offline copy of $\theta_{out}$.\\
    \textbf{6. Update Parameters:} 
    Combine losses and update $\theta_{rep}$:
    \[
    \mathcal{L}_{\text{rep}}(\theta_{rep}) = \frac{1}{N}\sum_{\mathbf{p}} ( \mathcal{L}_{\text{cnv}}(\theta_{rep}) + \lambda_1 \mathcal{L}_\text{{cst}}(\theta_{in}) + \lambda_2 \mathcal{L}_\text{{adv}}(\theta_{out}) ).
    \]
}

\BlankLine
\textbf{Training: Diffusion for Utterance Planning} \\
Freeze $\theta_{rep}$ and train $\theta_R$ using latent representations. \\
\For{each mini-batch of context $\mathbf{i}$ and segmented patches $\mathbf{P}$}{
    \textbf{1. Generate Corrupted Latent Representations:} \\
    Sample $\mathbf{Z}_t$ from $q(\mathbf{Z}_t | \mathbf{Z}_0)$. \\
    \textbf{2. Compute De-noising Loss:} 
    \[
        \mathcal{L}_\text{{noise}}(\theta_R) = \underset{\mathbf{P}, \mathbf{i}, t}{\mathbb{E}} \left\|  R(\mathbf{Z}_t, t, Enc_{ctx}(\mathbf{i}; \theta_{ctx})) - \mathbf{Z}_{t-1} \right\|^2_2.
    \]
    \textbf{3. Strengthen Guidance with Post-training:} 
    Freeze $\theta_{g}$ and $\theta_{dec}$ and compute:
    \[
        \hat{\mathbf{Z}}_{t} = R(\mathbf{Z}_{t+1}, t+1, Enc_{ctx}(\mathbf{i}; \theta_{ctx}))
    \]
    \[
    \mathcal{L}_{\text{rec}}(\theta_R) = \underset{\mathbf{P}, \mathbf{i}, t}{\mathbb{E}} \left\| g(\hat{\mathbf{Z}}_{t};\theta_g) - g({\mathbf{Z}}_{t};\theta_g) \right\|^2_2,
    \]
    \[
    \mathcal{L}_{\text{dec}}(\theta_R) = \underset{\mathbf{P}, \mathbf{i}, t}{\mathbb{E}} \left[ - \sum_{l=1}^p \log p(o_l | o_{<l}, \hat{\mathbf{Z}_t};\theta_{dec}) \right].
    \]
    \textbf{4. Update Parameters:} 
    Combine losses and update $\theta_R$:
    \[
    \mathcal{L}_{\text{diff}}(\theta_R) = \mathcal{L}_\text{{s2s}}(\theta_R) + \lambda_3 \mathcal{L}_{\text{rec}}(\theta_R) + \lambda_4 \mathcal{L}_{\text{dec}}(\theta_R)
    \]
}

\textbf{Return:} Trained model parameters $\theta_{rep}$ and $\theta_R$.
\end{algorithm*}

\begin{table*}
\centering
\footnotesize
\begin{tabular}{|c|p{0.25\textwidth}|p{0.25\textwidth}|p{0.25\textwidth}|}
\hline
\textbf{Noise Ratio} & \textbf{LD4LG} & \textbf{Adversarial} & \textbf{CSE + Adversarial} \\ \hline
 \multicolumn{4}{|c|}{\textbf{Original}: David noticed he had put on a lot of weight recently.} \\ \hline
0.00 & David noticed he had put on a lot of weight recently. & David noticed he had put on a lot of weight recently. & David noticed he had put on a lot of weight recently. \\ \hline
0.11 & David noticed he had put on a lot of weight recently. & David noticed he had put on a lot of weight recently. & David noticed he had put on a lot of weight recently. \\ \hline
0.22 & Amanda noticed he had put on a lot of weight recently. & David noticed he had put on a lot of weight recently. & David noticed he had put on a lot of weight recently. \\ \hline
0.33 & David he had to give up a thousand of cigarettes. & David noticed he had put on a lot of weight. & David noticed he had put on a lot of weight recently. \\ \hline
0.44 & he had put did on a lot of weight recently. & David a had put on a lot of weight. & David noticed he had put on a lot of weight recently. \\ \hline
0.56 & he noticed he had put on a lot of weight lately. & David -- he had to put on a lot of weight. & David realized he had put on a lot of weight recently. \\ \hline
0.67 & He expected he gotten ban on a lot of weight recently. & I had to give a dose of weight. & In mentioned he had put on, a lot of weight recently. \\ \hline
0.78 & It would smile a pa heard she had a sign for a little on & you have ad on a lot of food? & lization marche, 15 had jumped on to a lot of weight of a \\ \hline
0.89 & Some weeks contacted he's got, ration. & You are the threw in a defo. & The fat, hearing the of rounded. \\ \hline
1.00 & The sitter of Edward that she defended, and said. & , he, and the... & I thought, 10thed he would constantly ress, sanctwing of \\ \hline
\end{tabular}
\caption{Comparison of decoded text segment for varying noise ratios across three models: Vanilla, Noise Only, and CSE + Noise. The original sentence is listed separately for reference.}
\label{tab:decoded_text}
\end{table*}

\begin{table*}[!hbp]
\centering
\footnotesize
\begin{tabular}{p{0.2\textwidth}|p{0.7\textwidth}}
\toprule
\textbf{Title} & \textbf{Lilo's Walk} \\ \midrule
\textbf{Model} & \textbf{Generated Text} \\ \midrule
\texttt{Gold} & Sarah has a dog named Lilo. Sarah took Lilo for a walk. Lilo rolled around in dirt during his walk. Sarah had to take Lilo to the groomer. When Lilo came home from the groomer, he looked beautiful and clean. \\ \midrule
\texttt{LD4LG} & Lilo was about to take a walk to school. She saw no people on the grass. \textcolor{orange}{Lilio} was walking by the grass, she ran every grass. \textcolor{orange}{Llo} knew she was bored and walked to class. \\ \midrule
\texttt{Flan-T5} & Lilo wanted to go for a walk. Lilo's mom took her to the park. Lilo and her mom walked to the park. Lilo had a great time walking to the park. Lilo and her mom walked home together. \\ \midrule
\texttt{Ours} & Lilo was walking down the sidewalk for her dog. Lilo has not liked to walk across the woods. Lilo took him to walk in the park. Lilo's mom would walk with her every day. Lilo continued to walk again. \\ \bottomrule
\end{tabular}
\caption{Text quality comparison of generated outputs for the story "Lilo's Walk" in ROCStories. The table compares the outputs of LD4LG, Flan-T5, and our model against the gold standard.All model generates admissible story, however, LD4LG has minor word corruptions (Orange).}
\vspace{-3mm}
\label{tab:roc_comparison}
\end{table*}

\begin{table*}[ht]
\centering
\footnotesize
\begin{tabular}{p{0.2\textwidth}|p{0.7\textwidth}}
\toprule
\textbf{Title} & \textbf{Dialogue Summary: \#Person1\# invites Jimmy to go workout and persuades him into working out on arms and stomach.} \\ \midrule
\textbf{Model} & \textbf{Generated Text} \\ \midrule
\texttt{Gold} & \#Person1\#: Hey Jimmy. Let's go workout later today. \newline 
\#Person2\#: Sure. What time do you want to go? \newline 
\#Person1\#: How about at 3:30? \newline 
\#Person2\#: That sounds good. Today we work on Legs and forearm. \newline 
\#Person1\#: Hey. I just played basketball earlier, so my legs are a little sore. Let's work out on arms and stomach today. \newline 
\#Person2\#: I'm on a weekly schedule. You're messing everything up. \newline 
\#Person1\#: C'mon. We're only switching two days. You can do legs on Friday. \newline 
\#Person2\#: Aright. I'll meet you at the gym at 3:30 then. \\ \midrule
\texttt{Flan-T5} & \#Person1\#: Hey, Jimmy, why don't you come work out with me?
\newline \#Person2\#: What are you talking about? \newline \#Person1\#: I'm talking about working out on your arms and your stomach. \newline \textcolor{Brown}{\#Person2\#: What are you talking about? \newline \#Person1\#: I mean, you lift weights on your arms and you lift weights on your stomach. \newline \#Person2\#: That's not working. \newline \#Person1\#: But you lift weights on your arms and you lift weights on your stomach. \newline \#Person2\#: That's not working. \newline \#Person1\#: But you lift weights on your arms and you lift weights on your stomach. \newline \#Person2\#: That's not working. \newline 
\#Person1\#: But you lift weights on your arms and you lift weights on your stomach. \newline 
\#Person2\#: That's not working. (repeated 7 times more)} \\ \midrule
\texttt{LD4LG} & \#Person1\#: Hi, Jimmy. We're going to do some exercises and arm exercises on this weekend. Would you like to join us? \newline 
\textcolor{orange}{\#PEerson2\#}: Yeah, that sounds good for us. \newline 
\textcolor{orange}{\#PEPerson3\#}: Yeah, working on arms with arms, but I've thought I'm going to start on my stomach. \newline 
\textcolor{orange}{\#EPerson4\#}: Working??.. I don't think I just have to hold up my weight. \newline 
\textcolor{orange}{\#EPerson3\#}: Why not go on your arms and arms and then put arms on your stomach. Then you can work for for any time. \\ \midrule
\texttt{Ours} & \#Person1\#: Hey Jimmy, how's it going? \newline 
\#Person2\#: I'm still not sure, physically. \newline 
\#Person1\#: So, what's going on? \newline 
\#Person2\#: I'm just not feeling up to it. \newline 
\#Person1\#: Come on, let's get moving. \newline 
\#Person2\#: What do you mean? \newline 
\#Person1\#: Well, we could work on your arms and stomach. \newline 
\#Person2\#: Oh, no thank you. I don't know it as well. \newline 
\#Person1\#: It's not that hard, I'll show you. \newline 
\#Person2\#: Okay, I'll try. \newline 
\#Person1\#: That's the spirit! \\ 
\bottomrule
\end{tabular}
\caption{Text quality comparison of generated dialogues for the summary: "\#Person1\# invites Jimmy to go workout and persuades him into working out on arms and stomach" in DialogSum. The table compares the outputs of LD4LG, Flan-T5, and our model against the gold standard. Flan-T5 generates repetitive text (Brown), while LD4LG sometimes corrupts usernames (Orange). }
\vspace{-3mm}
\label{tab:dialsum_comparison}
\end{table*}

\begin{table*}[ht]
\centering
\footnotesize
\begin{tabular}{p{0.2\textwidth}|p{0.7\textwidth}}
\toprule
\textbf{Meta-Info} & \textbf{Users: [USER\_SYS] [USER\_0] [USER\_1] Cards: 4 9 V U}\\ \midrule
\textbf{Previous Turns} & [USER\_0]HI! I went with flipping the card with the letter U \\ \midrule
\textbf{Model} & \textbf{Generated Text} \\ \midrule
\texttt{Gold} & [USER\_1]I selected U and 9 [USER\_1]You need to select 9 also, the other side might be a vowel [USER\_0]I completely overlooked that you could select 2 cards. I agree both should be selected [USER\_0] [PICK] \textbf{U} \textbf{9} [USER\_1]the other two cards are irrelevant [USER\_0]yes i agree  [USER\_1]Are you ready to submit the HIT? [USER\_0]yes, i submitted my final answer [USER\_1] [PICK] \textbf{U} \textbf{9} [USER\_SYS][FINISHED] \\ \midrule
\texttt{LD4LG} & [USER\_0] Hello [USER\_0] I picked the letter "U" and the number "9". I chose u and \textbf{ 9} because I think those are the ones you need to check to see if the rule is correct? I didn't think about it that way. I was thinking that it should be 2 cards with a vowel and a number on the other side. I also did not see any reason why I chose the same card.  [USER\_1] I picked U, 9 and [PICK] \textbf{ 4} \textbf{ U}. Why did you pick the letter \textcolor{red}{\textbf{ 8}} and u? I thought it was 2 cards that would verify the rule. So I picked 2 cards and ur correct answer  \\ \midrule
\texttt{Flan-T5} & [USER\_0] Hi, it was nothing like this. [PICK] \textbf{U} \textbf{9} [USER\_0] I have chosen the cards to flip over that one. it does say an even number would have to have a vowel and a consonant. [USER\_0] You think that would work well to prove either of the 2 cards? [USER\_0] I did what you said [USER\_1] you should flip them to see if there's a vowel or consonant on the side (apart from the number) [USER\_0] hmmmmm i could imagine choosing them all because the rule needs to be followed. [USER\_1] maybe we should keep to u and 9 then? [USER\_1] alright I'm gonna submit those two I guess [USER\_1] I don't think we need to do all four as its not a vowel so I don't think we need to see it because the rule doesn't state that consonants can't have an even number or vowel [USER\_0] Oh really? lol [USER\_1] so \textcolor{brown}{u and 9?} [USER\_1] yea it is a vowel and a consonant [USER\_1] we're testing the rule so turning them all would test the rule [USER\_0] well guess we'll see what happens when we're done [USER\_1] it says to ONLY flip the cards that have letters. i think to check the rule that cards with vowels have an even number so that's the only one you need to flip [USER\_0] [PICK] \textbf{U} [USER\_0] how do we know that the rule doesn't actually apply to all evens [USER\_1] we're testing the rule to see if there is a vowel on the other side [USER\_1] but it doesn't specify if there's a non vowel on the other side [USER\_0] the task asks if there is a number in each of the 3 cards [USER\_0] oh yeah so just the card with the vowel would test it [USER\_0] yeah :) [USER\_0] but then we have to test the other two cards [USER\_0][USER\_1] hmmm lmao! should we flip them all then? [USER\_0] well yeah, mine seems reasonable. [USER\_0] tiger sorry im confused lol lol [USER\_0] u and 9 maybe too [USER\_2] ok so maybe not all 3? [USER\_0] yeah \textcolor{brown}{u and 9?} [USER\_0] thats what im going with emuerk? [USER\_0] yea I see where you are coming from lol [USER\_1] we're trying to test the rule, but its not necessarily involving vowels or consonants [USER\_1] so just \textcolor{brown}{u and 9?} [USER\_1] yea tiger, we have to participate anyways [USER\_0] yeah yea [USER\_0] so \textcolor{brown}{u and 9?} [USER\_0] yea [USER\_2] if u is a vowel, it explains the rule about vowels. if v is an odd number it doesn't mean the rule is wrong [USER\_0] \textcolor{brown}{u and 9?} [USER\_2] yea \textcolor{brown}{u and 9?} [USER\_0] yeah both would show the rule [USER\_2] [PICK] \textbf{U} \textbf{9} [USER\_2] yeah u and 9 is my thinking too. [USER\_0] same [USER\_1] so what do we do now [USER\_0] submit \textcolor{brown}{u and 9?} [USER\_1] [PICK] \textbf{U} \textbf{9} [USER\_1] now im thinking \textcolor{brown}{u and 9?} \textcolor{red}{[USER\_2]} [PICK] \textbf{U} \textbf{9} \textcolor{red}{[USER\_2]} [PICK] \textbf{U} \textbf{9} [USER\_SYS] [FINISHED] \\ \midrule
\texttt{Ours} & [USER\_1] I have the same opinion. [USER\_0] What do you think? [USER\_0] And I would also need to flip the \textbf{ 9} to make sure there's not a vowel on the back [USER\_0] I have the same number [USER\_0] why \textbf{ 4}? [USER\_0] I think we need to flip the \textbf{ 4} as well to see if there is a vowel [USER\_0] Oh, that makes sense [USER\_0] and flip the \textbf{ U}, to see if there is an even number [USER\_0] I kind of like this good idea [USER\_1] [PICK] \textbf{ U} \textbf{ 4} [USER\_0] [PICK] \textbf{ U} \textbf{ 4} [USER\_SYS][FINISHED] \\ \bottomrule
\end{tabular}
\caption{Text quality comparison of generated continuations for a DeliData example. The table compares the outputs of LD4LG, Flan-T5, and our model against the gold standard. LD4LG hallucinated a card choice (Red). Flan-T5 generates unnatural repetitive outputs (Brown) and hallucinate an extra user (Red).}
\vspace{-3mm}
\label{tab:delidata_comparison}
\end{table*}

\begin{table*}[!hbp]
\centering
\footnotesize
\begin{tabular}{p{0.2\textwidth}|p{0.7\textwidth}}
\toprule
\textbf{Input} & \textbf{Answer the following question by reasoning step by step: antarctic ocean is named after a continent. What's another ocean that's named after a landmass? Choices: antarctic, atlantic, pacific, indian ocean, pacific.} \\ \midrule
\textbf{Model} & \textbf{Generated Text} \\ \midrule
\texttt{Gold} & India is a name of a country which is a land mass. So Indian ocean is named after a landmass. Antarctic is always given in the question and the other options are not names of land masses. The answer is \textcolor{ForestGreen}{indian ocean}.\\ \midrule
\texttt{Flan-T5 Base} & atlantic ocean is named after a continent. atlantic ocean is named after a landmass. Therefore, the final answer is \textcolor{red}{atlantic}. \\ \midrule
\texttt{LD4LG} & The atlantic ocean is named for antarcam. So atlantic is named an antarcha ocean. All the other options are not named another ocean named for landmass. The answer is \textcolor{red}{atlantic}. \\ \midrule
\texttt{Ours} & The antarcic ocean is named to pacific because it is named another an landmass. So the antico is named by atlantic ocean. All the other options are not named by which ant ocean is called another after landmasses. The answer is \textcolor{red}{atlantic}. \\ \midrule
\texttt{Flan-T5 Large} & Pacific Ocean is named after pacific. Pacific Ocean is named after pacific. Therefore, the final answer is \textcolor{red}{pacific}. \\ \bottomrule
\end{tabular}
\caption{Models' Predictions on ECQA: All diffusion models gives a wrong answer that is consistent with the knowledge of their Flan-T5 Base backbone.}
\vspace{-3mm}
\label{tab:ecqa_all_wrong}
\end{table*}

\begin{table*}[!hbp]
\centering
\footnotesize
\begin{tabular}{p{0.2\textwidth}|p{0.7\textwidth}}
\toprule
\textbf{Input} & \textbf{Answer the following question by reasoning step by step: If someone brings candy to a group of children what will happen? Choices: listen to music, spring into action, wave goodbye, play chess, experience fear.} \\ \midrule
\textbf{Model} & \textbf{Generated Text} \\ \midrule
\texttt{Gold} &  Springing into action means to become active unlike playing chess and experiencing fear. Candies make children spring into action and not listen to music or wave goodbye. The answer is \textcolor{ForestGreen}{spring into action}.\\ \midrule
\texttt{Flan-T5 Base} & Spring into action is the act of bringing something to a group of children. If someone brings candy to a group of children, spring into action will happen. Therefore, the final answer is \textcolor{ForestGreen}{spring into action}. \\ \midrule
\texttt{LD4LG} & Springing into action leads to spring into action and not play chess or wave fear. Springing onto action which leads to Spring into music unlike listening to music and experiencing fear. The answer is \textcolor{brown}{
spring into fear}. \\ \midrule
\texttt{Ours} & Spring into action is an activity unlike listening to music, play chess. Candy makes a group of children springing into action. The answer is \textcolor{ForestGreen}{spring into action}. \\ \midrule
\texttt{Flan-T5 Large} & Spring into action is to get excited. Candy is a good thing. Children get excited when someone brings candy to them. Therefore, the final answer is \textcolor{ForestGreen}{spring into action}. \\ \bottomrule
\end{tabular}
\caption{Models' Predictions on ECQA: LD4LG initially mentions the correct answer, but predicts a non-existent answer by blending choices ``spring into action'' and ``experience fear'' together.}
\vspace{-3mm}
\label{tab:ecqa_ld4lg_wrong}
\end{table*}

\end{document}